%% file: main.tex
\documentclass[]{template}

\usepackage[utf8]{inputenc}             
\usepackage[T1]{fontenc}                
\usepackage{url}                        
\usepackage{booktabs}                   
\usepackage{multirow}
\usepackage{colortbl}
\usepackage{multicol}
\usepackage{amsfonts}                   
\usepackage{nicefrac}                   
\usepackage{microtype}                  
\usepackage[dvipsnames]{xcolor}         

\usepackage{latexsym}

\usepackage{graphicx}
\usepackage{float}
\usepackage{subcaption}
\usepackage{wrapfig}
\usepackage{lipsum}

\usepackage{bm}

\usepackage{tabularx} 
\usepackage{ragged2e} 
\newcolumntype{L}{>{\RaggedRight\hangafter=1\hangindent=0em}X}

\usepackage{enumitem}

\usepackage{amsmath}
\usepackage{amssymb}
\usepackage{mathtools}
\usepackage{amsthm}

\setboolean{logo}{true}    

\usepackage[linesnumbered,ruled,vlined]{algorithm2e}

\hypersetup{
    colorlinks=true,
    linkcolor=red,
    citecolor=Cerulean,
    filecolor=magenta,      
    urlcolor=magenta,
}

\usepackage[capitalize,noabbrev]{cleveref}
\crefname{section}{§}{§§}
\Crefname{section}{§}{§§}

\usepackage{calligra}
\DeclareMathAlphabet{\mathcalligra}{T1}{calligra}{m}{n}

\usepackage{pifont}

\theoremstyle{plain}

\theoremstyle{definition}

\theoremstyle{remark}

\renewcommand{\paragraph}[1]{\vspace{1mm}\noindent\textbf{#1}}

\DeclareCaptionLabelFormat{cont}{#1~#2\alph{ContinuedFloat}}
\usepackage[most]{tcolorbox}
\tcbset{
  promptbox/.style={
    top=10pt,
    colback=lightgray!20,
    colframe=Black,
    colbacktitle=NavyBlue,
    enhanced,
    center,
    attach boxed title to top center={yshift=-0.1in,xshift=0.0in},
    boxed title style={boxrule=0pt,colframe=white,},
  }
}
\newtcolorbox{promptbox}[2][]{promptbox, title=#2,#1}
\tcbset{
  takeawaybox/.style={
    top=10pt,
    colback=lightgray!20,
    colframe=Black,
    colbacktitle=BurntOrange,
    enhanced,
    center,
    attach boxed title to top center={yshift=-0.1in,xshift=0.0in},
    boxed title style={boxrule=0pt,colframe=white,},
  }
}
\newtcolorbox{takeawaybox}[2][]{takeawaybox, title=#2,#1}
\tcbset{
  observationbox/.style={
    top=10pt,
    colback=lightgray!20,
    colframe=Black,
    colbacktitle=YellowGreen,
    enhanced,
    center,
    attach boxed title to top center={yshift=-0.1in,xshift=0.0in},
    boxed title style={boxrule=0pt,colframe=white,},
  }
}
\newtcolorbox{observationbox}[2][]{observationbox, title=#2,#1}

\usepackage{xspace}

\newcommand\blfootnote[1]{%
  \begingroup
  \renewcommand\thefootnote{}\footnote{#1}%
  \addtocounter{footnote}{-1}%
  \endgroup
}

\usepackage{CJK}

\title{OpenCompass: A Universal Evaluation Platform for Large Language Models}

\author{OpenCompass Team, Shanghai AI Laboratory}

\input{sections/0.abstract}

\begin{document}

\blfootnote{$*$ Code is at \url{https://github.com/open-compass/opencompass}}

\maketitle

\input{sections/1.intro}
\input{sections/2.pre}
\input{sections/3.arch}
\input{sections/4.benchmark}
\input{sections/5.future}

\clearpage
\bibliographystyle{plain}
\bibliography{refs}


\clearpage
\appendix
\input{sections/6.appendix}



\end{document}

%% file: sections/0.abstract.tex
\begin{abstract}

In recent years, the field of artificial intelligence has undergone a paradigm shift from task-specific small-scale models to general-purpose large language models (LLMs). With the rapid iteration of LLMs, objective, quantitative, and comprehensive evaluation of their capabilities has become a critical link in advancing technological development. Currently, the mainstream static benchmark dataset-based evaluation methods face challenges such as the diversity of task types, inconsistent evaluation criteria, and fragmentation of data and processing workflows, making it difficult to efficiently conduct cross-domain and large-scale model evaluation. 
To address the aforementioned issues, this paper proposes and open-sources OpenCompass, a one-stop, scalable, and high-concurrency-supported general-purpose LLM evaluation platform. Adhering to the design philosophy of modularization and component decoupling, the platform boasts three core advantages: high compatibility, flexibility, and high concurrency.
The core architecture of OpenCompass comprises five key components: the Configuration System, Task Partitioning Module, Execution and Scheduling Module, Task Execution Unit, and Result Visualization Module. Its workflow provides rule-based, LLM-as-a-Judge, and cascaded evaluators to adapt to the requirements of different task scenarios. Supporting mainstream benchmark datasets across multiple domains, including knowledge, reasoning, computation, science, language, code, etc., the platform offers a unified and efficient LLM evaluation tool for both academia and industry, facilitating the accurate identification of strengths and weaknesses of LLMs as well as their subsequent optimization.

\end{abstract}

%% file: sections/1.intro.tex
\section{Introduction}

In recent years, the field of artificial intelligence has undergone a paradigm shift from task-specific small-scale models to general-purpose large language models (LLMs). 
Compared with traditional models that are small in parameter size and often require fine-tuning for specific downstream tasks, large language models exhibit remarkable generalization and emergent abilities through pre-training on massive corpora. 
This scale effect enables LLMs not only to tackle tasks on which they have not been explicitly trained, but also to possess powerful in-context learning capabilities. 
Although current large language models are rapidly evolving toward a multimodal orientation, integrating multiple perceptual modalities such as vision and audition, text processing capability — as the core carrier for logical reasoning, knowledge storage, and instruction interaction — remains a crucial cornerstone for the construction of high-level machine intelligence.

With the rapid iteration of LLMs, the objective and quantitative evaluation of their capabilities has become a critical link in advancing technological development. 
Constructing a comprehensive capability evaluation system typically encompasses multiple dimensions, including general knowledge reserve, language understanding and generation, complex instruction following, logical reasoning, and professional domain-specific capabilities in vertical fields (e.g., medicine, law, coding, etc.). 
Through the evaluation of general and domain-specific capabilities based on the aforementioned multi-dimensional framework, researchers can accurately identify the strengths and weaknesses of models, thereby providing data support for subsequent model optimization.

Currently, the academic and industrial communities primarily adopt static benchmark dataset-based methods for LLM evaluation. 
This approach quantifies the performance of models on specific tasks by examining its answer accuracy or generation quality across various standard test suites (e.g., MMLU \cite{mmlu}, HLE \cite{hle}, GPQA \cite{gpqa}.). 
However, a holistic assessment of the overall quality of LLMs often requires testing across dozens or even hundreds of benchmarks covering diverse domains and difficulty levels.

Due to the diversity of task types and the variability of evaluation criteria, unified management of these scattered tests has become an extremely arduous task. 
Specifically, these differences mainly include: inconsistent evaluation methods for model outputs across benchmarks (e.g., regular expression matching, ROUGE scoring, model-based scoring), high sensitivity of results to prompt engineering, and highly fragmented preprocessing and postprocessing of data and model outputs among different benchmarks.

To address the fragmentation and standardization challenges faced in the aforementioned evaluation process, we have developed and open-sourced OpenCompass — a one-stop, scalable, and high-concurrency-supported general-purpose LLM evaluation platform. 
It aims to provide a unified and efficient evaluation tool for both academia and industry.

Specifically, OpenCompass exhibits the following prominent advantages:

1. \textbf{Comprehensive and Extensive Support for Models and Datasets}: OpenCompass supports the evaluation of 100+ mainstream datasets, covering dimensions including disciplinary knowledge, linguistic competence, factual knowledge, comprehension ability, reasoning capability, and safety compliance, as well as specialized capabilities such as long-text processing and code generation. Meanwhile, it is compatible with multiple types of large models, including HuggingFace-based models, accelerated inference engine models, API models, and various custom models.

2. \textbf{Flexible and Diversified Evaluation Strategies}: OpenCompass integrates multiple evaluation paradigms under subjective evaluation and objective evaluation. It also supports the customization of evaluation logic, data preprocessing, and postprocessing methods across different benchmarks.

3. \textbf{Efficient Concurrent Evaluation Architecture}: To address the challenges of high computational overhead and long response times in large model evaluation, OpenCompass is designed with an efficient task partitioning mechanism. It supports large-scale, cluster-based parallel inference, significantly reducing the time required to complete full-scale evaluations on massive datasets.

%% file: sections/2.pre.tex
\section{Preliminary}

LLM evaluation refers to the systematic assessment of large language models across key dimensions such as knowledge breadth, reasoning ability, generalization performance, and safety through a series of standardized testing methods. In existing evaluation systems, the evaluated objects are typically categorized into two types: Base Models, i.e., models pretrained solely on massive corpora with text continuation capabilities; and Chat Models, i.e., models that have undergone Supervised Fine-Tuning (SFT) or Reinforcement Learning from Human Feedback (RLHF), equipped with the abilities to follow complex instructions and conduct multi-turn interactions. For different types of models and tasks, evaluation methods are mainly divided into the following two categories:

\textbf{Objective Evaluation}: Primarily targeted at tasks with definitive gold standard answers, such as multiple-choice questions, mathematical question answering, or code generation. This type of evaluation achieves efficient, low-cost, and fully reproducible automated assessment by comparing model outputs with gold standard answers, utilizing quantitative metrics including Accuracy, Exact Match, or F1 score.

\textbf{Subjective Evaluation}: Mainly applied to open-ended generation tasks, such as creative writing, logical debate, or complex instruction response. Due to the lack of a unique gold standard answer, this type of evaluation typically relies on the LLM-as-a-Judge paradigm to perform multi-dimensional quality scoring on generated content across dimensions like coherence, usability, and instruction adherence.

Typically, a complete evaluation task executed on static benchmarks follows a typical pipeline consisting of three phases: Data Preprocessing, Model Inference, and Result Evaluation.

\textbf{Data Preprocessing}: Reads samples from raw datasets, performs format standardization and field alignment, and converts them into the input format required by the model.

\textbf{Model Inference}: Constructs prompts and inputs them into the LLM for forward propagation to generate responses. Inference paradigms vary across different model types: base models generally adopt perplexity-based methods, while chat models commonly use generation-based approaches. Additionally, to address the efficiency challenges of large-scale evaluation, this phase typically involves batch inference and the scheduling of distributed computing resources.

\textbf{Result Evaluation}: As the final link in the evaluation pipeline, it aims to convert the raw outputs into quantifiable scores. This phase involves post-processing of model outputs, answer extraction, and scoring.

%% file: sections/3.arch.tex
\section{Architecture}

\begin{figure}[t!]
    \centering
    \includegraphics[width=0.9\linewidth]{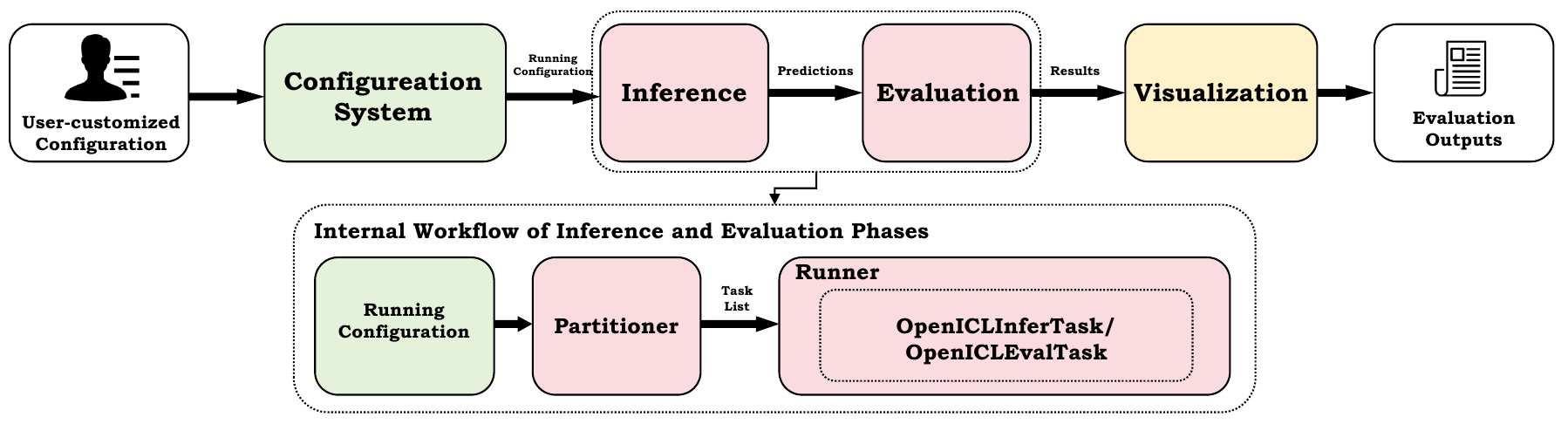}
    \caption{Overview of the OpenCompass architecture}
    \label{fig:arch}
\end{figure}

The project architecture of OpenCompass adheres to the design philosophy of modularization and component decoupling, aiming to construct a general-purpose evaluation framework with high efficiency, high compatibility, and high scalability. As illustrated in Figure \ref{fig:arch}, the overall architecture of OpenCompass adopts a layered logical architecture, comprising five core components: the Configuration System; Partitioners (task partitioning modules); Runners (execution backends and scheduling modules); Tasks (task execution units); and Summarizers (result visualization modules).

The evaluation workflow of OpenCompass is abstracted into four standard stages: the Configuration Construction Stage, Inference Stage, Evaluation Stage, and Visualization Stage. Its specific execution logic is as follows:

First, in the Configuration Construction Stage, users define specific evaluation configurations, specifying model architectures, dataset specifications, and relevant evaluation parameters.
Second, entering the Inference Stage, the system instantiates the evaluation dataset according to the configurations and drives the model to generate prediction results.
Subsequently, in the Evaluation Stage, the system validates the inference outputs and computes evaluation metrics. Finally, in the Visualization Stage, persistent storage and visualization presentation of the evaluation results are performed.
Notably, both the Inference and Evaluation Stages adhere to a unified task execution paradigm: first, the Partitioner decomposes the overall evaluation load into multiple independent atomic tasks based on preset strategies; further, the Runner is responsible for computing resource management and distributes these tasks in parallel to designated computing backends — local processes or cluster jobs — for execution. Subsequent sections will elaborate on the implementation details of the aforementioned workflow in depth.

\subsection{Configuration System}

\begin{figure}[t!]
    \centering
    \includegraphics[width=0.9\linewidth]{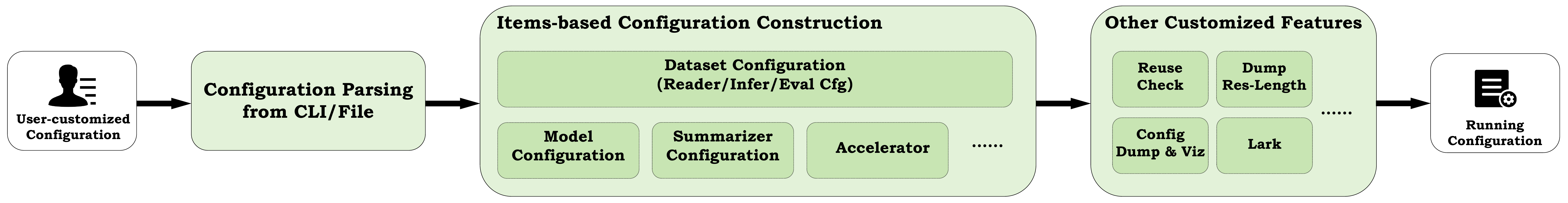}
    \caption{Architecture of the Configuration System}
    \label{fig:arch_cfg}
\end{figure}

The lifecycle of the program commences with the configuration entry point. OpenCompass accepts inputs from Python configuration files or the Command Line Interface (CLI), where users define model parameters, dataset specifications, and evaluation strategies. Built on MMEngine \cite{mmengine}, the system parses, aligns, and instantiates the input heterogeneous configurations, constructs a global unified evaluation configuration object, and completes consistency checks for the working directory and data, as illustrated in Figure \ref{fig:arch_cfg} . The core configuration content primarily includes dataset configurations and model configurations.

To define a complete dataset configuration, users are required to construct data loading configurations, inference configurations, and evaluation configurations. Notably, the configuration system of OpenCompass, built on OpenICL \cite{openicl}, supports a highly flexible prompt construction mechanism to adapt to the in-context learning requirements of diverse tasks. This mechanism mainly encompasses three types: 
1) \textbf{Few-shot prompts}: Implemented based on the Retriever module. By configuring retrieval strategies, specified contextual examples are selected from the indexed dataset and concatenated with test samples to form complete prompts containing examples. 
2) \textbf{Zero-shot prompts}: Directly constructed via the Prompt Template module, consisting only of task descriptions and test samples without additional contextual examples. They are suitable for task scenarios where models possess zero-shot generalization capabilities. 
3) \textbf{Rich prompt structures}: Also constructed through the Prompt Template module and mapped to ChatML templates, enabling support for static multi-turn contextual prompts.

\subsection{Partitioning and Parallelism}
\begin{figure}[t!]
    \centering
    \includegraphics[width=0.9\linewidth]{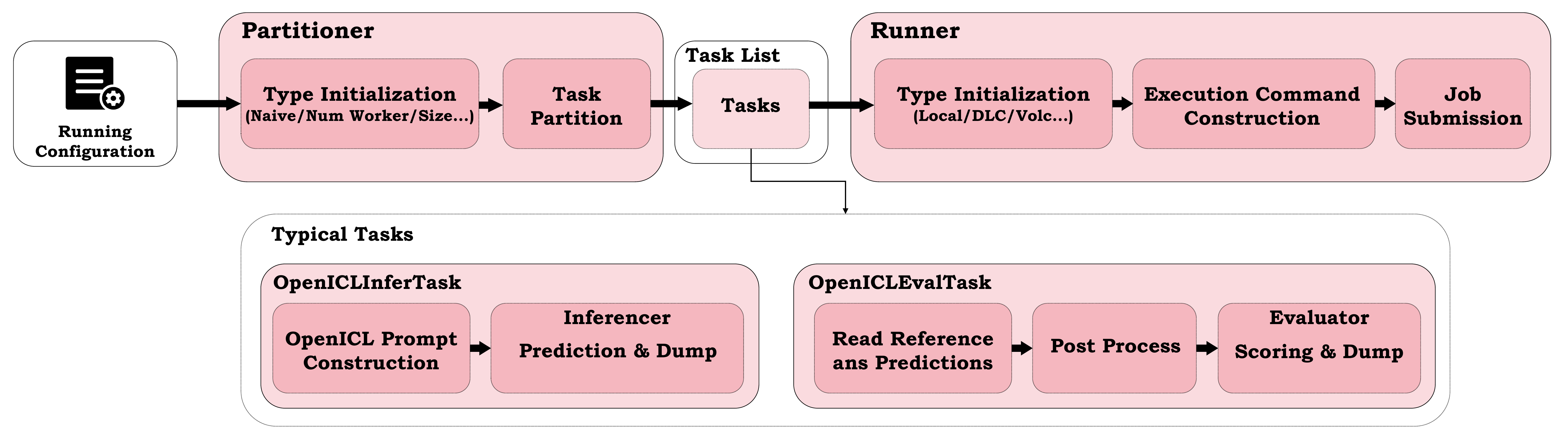}
    \caption{Architecture of Partitioner, Runner and Tasks}
    \label{fig:arch_task}
\end{figure}

The computationally intensive nature of large language model inference and the large-scale characteristic of evaluation datasets result in the time complexity of serially executing complete evaluation tasks exhibiting a significant linear increase with the task scale, making it difficult to meet the efficiency and timeliness requirements in large-scale model evaluation scenarios. To address this challenge, OpenCompass deploys two core components, Partitioner and Runner, in the core workflows of the inference and evaluation stages for efficient resource scheduling, as shown in Figure \ref{fig:arch_task}. Their primary goal is to decompose computationally expensive complete evaluation tasks into a number of independent and mutually non-dependent subtasks, thereby fully tapping into the potential of cluster computing resources through distributed parallel execution and drastically reducing the evaluation cycle.

After completing the parsing and alignment of evaluation task configurations, the system first maps the evaluation requirements involving multiple models and datasets to the Cartesian product set of model-dataset combinations—i.e., enumerating all pairwise combinations of user-specified model lists and dataset lists. Subsequently, the Partitioner divides the tasks based on user-defined partitioning criteria, such as the number of workers, dataset size, or naive equal partitioning. It then constructs independent output file paths for each subtask and packages all subtasks into a structured task list, which is passed to the Runner. As the atomic units of evaluation execution, these subtasks can be run in parallel through subsequent threading and distributed scheduling, thereby minimizing the bubble ratio in the pipeline and significantly improving computational resource utilization.

In heterogeneous multi-machine multi-GPU cluster environments, the parallel execution of large-scale evaluation tasks highly relies on the job allocation and resource scheduling capabilities of the underlying cluster management systems. However, significant heterogeneity exists in the scheduling interfaces and resource models of different cluster management systems, which greatly increases the complexity of cross-environment deployment of evaluation tasks. In the architecture of OpenCompass, the final task allocation, resource coordination, and cluster interaction logic are uniformly undertaken by the Runner component. Serving as an adaptive abstraction layer between the evaluation task execution logic and the underlying cluster management systems, it can shield the heterogeneity of cluster environments and realize standardized and efficient scheduling of evaluation tasks across different clusters.

Specifically, the workflow of the Runner can be decomposed into the following stages: First, the Runner performs cluster environment adaptation and initialization. OpenCompass provides built-in support for various cluster environments, including local standalone environments, Alibaba Cloud Data Intelligence Platform, Volcengine Cloud Cluster, etc. Subsequently, the Runner conducts standardized conversion and submission of task instructions—parsing the structured task list and computational resource requirements received from the Partitioner, and converting them into execution instructions compliant with the target cluster management system (in debugging scenarios, submissions are made in a thread-based manner). Additionally, the Runner is responsible for the global management of the task lifecycle: it tracks the execution status of each subtask in real time, triggers a configurable retry mechanism for failed tasks, and collaboratively manages subtask logs and output files.

\subsection{Task}

The structured task list constructed by the Partitioner serves as the core input carrier for the Runner task scheduling. Each Task within the task list is an indivisible minimal execution unit in the evaluation pipeline, independently carrying the complete execution information and context of a single-batch task. OpenCompass primarily supports two types of tasks: OpenICLInferTask and OpenICLEvalTask, which serve as the execution carriers for the inference stage and evaluation stage, respectively. Both tasks encompass functions of initialization, module execution, and log/output content management.

In the execution flow of OpenICLInferTask, model instantiation is first completed based on predefined model configurations. OpenCompass supports multi-modal model integration: it includes not only model integration via API calls but also native model deployment based on Hugging Face. Meanwhile, it can replace Huggingface native solution with inference acceleration frameworks such as vLLM \cite{vllm} and LMDeploy \cite{lmdeploy} to implement function-level model deployment, thereby improving inference efficiency. For raw datasets, samples undergo a pipeline processing flow consisting of the Retriever component and Template component to form the final model inputs. The Inferencer component is responsible for the unified scheduling of models and inputs, driving the model to perform inference computations to generate predictions, which are then dumped.

OpenICLEvalTask mainly focuses on the instantiation and execution of the Evaluator component. Specifically, this task first invokes the reference labels built into the dataset and loads the predictions output by the corresponding inference task. Subsequently, based on the user-configured PostProcessor component, standardized postprocessing operations such as format normalization and redundant information filtering are performed on the reference labels and predictions. Finally, the processed data from both ends is passed to the Evaluator component, which computes scores for the predictions according to preconfigured evaluation metrics (e.g., Accuracy, ROUGE, BLEU) and generates structured evaluation results.

\subsection{Evaluators} 
OpenCompass incorporates a variety of built-in Evaluator templates, which can adapt to the evaluation characteristics of different tasks and cover multiple evaluation scenarios such as objective rule-based assessment, subjective judgment, and hybrid cascading. The specific types and implementations are as follows:

\textbf{Rule-Based Evaluator} implements lightweight and efficient objective evaluation based on predefined logical rules, classic NLP evaluation metrics, or domain-specific parsing logic, without requiring additional model dependencies. It is suitable for tasks with clear result formats and quantifiable evaluation logic. Currently, OpenCompass supports three subtypes: 
1) \textit{Option Extraction Evaluator}: Designed for multiple-choice tasks, it extracts standardized options from the target model's generated text through rules such as string matching and regular expression parsing; 
2) \textit{Content Regex Evaluator}: Applicable to entity extraction and format-constrained tasks, it extracts target content for evaluation using predefined regular expressions. It also includes MathEvaluator, which verifies result consistency via a LaTeX formula parser; 
3) \textit{Other Classic NLP Metric Calculators}: Including BLEU, ROUGE, AUC-ROC, F1 Score, and other widely used metrics.

\textbf{LLM-as-a-Judge Evaluator} invokes an LLM serving as the Judge Model, reuses inference task logic, and guides it to complete evaluation through structured prompts. It is suitable for scenarios that are difficult to cover by rule-based methods, mainly including: 
1) \textit{Complex Objective Task Scenarios}: Adopted when answers or model responses are relatively complex and difficult to judge via rules, guiding the Judge Model to perform qualitative or quantitative scoring on model outputs; 
2) \textit{Subjective Task Scenarios}: For highly subjective tasks such as open-ended questions and creative collaboration, it takes the output of a representative and state-of-the-art closed-source LLM as the reference frame. Through prompts, the Judge Model is guided to perform relative quality ranking or multi-dimensional scoring of the target model's output against the reference output, considering dimensions such as relevance, fluency, logic, and innovativeness.

\textbf{Cascade Evaluator} combines the advantages of rule-based evaluation and LLM-based evaluation, offering two operating modes: 
1) \textit{Cascaded Mode}: Adopts a cascaded framework of rule-based pre-filtering and LLM Judge evaluation. In the first phase, a lightweight rule-based Evaluator performs preliminary screening of prediction results, quickly filtering out samples determined to be correct by rules. Samples judged as incorrect are passed to the LLM Judge Evaluator for further quality assessment. This approach maintains accuracy while reducing reliance on LLM judgment, thereby lowering evaluation costs and time; 
2) \textit{Parallel Mode}: Both the rule-based evaluator and the LLM judge evaluate all samples simultaneously. If either evaluator deems a sample correct, the sample is considered correct. This mode improves evaluation tolerance but may incur higher costs, as all samples require LLM evaluation. In both modes, the Evaluator outputs rule-based evaluation accuracy, LLM-based evaluation accuracy, and combined accuracy. It is particularly suitable for scenarios requiring a balance between evaluation cost and accuracy, cases where rule-based evaluators are available but incomplete, and evaluation tasks that demand precise judgment of edge cases, but mandates that the evaluation task adopt metrics with objective correctness criteria.

\subsection{Visualization}
After the full execution of the inference and evaluation stages, the final phase of OpenCompass evaluation workflow is the Visualization Stage. This stage is dominated by the core logic of the Summarizer component, which is responsible for completing the aggregation of evaluation results, calculation of subjective result metrics, and multi-form visualization presentation.

Specifically, the Visualization Stage supports refined and customized configuration of visualization results by users. Taking the predefined ``abbr'' field in the dataset configuration as the unique identifier of the dataset, users can specify the dimensions of evaluation metrics to be displayed. The Summarizer completes the associative mapping between evaluation results and dataset identifiers based on this configuration, and ultimately generates a customized visualization summary report that meets user requirements.

For scenarios involving composite datasets with sub-datasets, OpenCompass completes task splitting and parallel execution based on the unique ``abbr'' identifiers of each sub-dataset during the initialization phase of the evaluation task. In the visualization phase, sub-datasets can be grouped into the same aggregation unit through summary groups, and unified metrics are calculated in accordance with user-defined aggregation rules.

%% file: sections/4.benchmark.tex
\section{Benchmarks}

To date, OpenCompass supports the evaluation of over 100 datasets, covering mainstream benchmark suites across key domains such as knowledge, reasoning, computation, science, language, code, and long-text processing. The following sections list the datasets currently supported by the project. For more supported datasets, please refer to \url{https://opencompass.readthedocs.io/en/latest/dataset_statistics.html}.
Furthermore, we present evaluation results for most mainstream models in the current LLM community on several popular, publicly available datasets, sourced from the OpenCompass Academic Leaderboard. See Appendix A for details.

\subsection{Knowledge}

\textbf{MMLU \cite{mmlu}}: A benchmark for comprehensively evaluating the multi-domain knowledge reserve and basic reasoning capabilities of large language models. It covers multiple-choice questions across 57 different subject areas (including humanities, social sciences, natural sciences, technology, etc.), with the core objective of assessing the model's mastery and basic application of foundational factual knowledge and concepts in various domains. 

\textbf{GPQA \cite{gpqa}}: A benchmark for evaluating the general factual knowledge reserve of large language models. Its questions cover cross-domain general knowledge such as science, history, literature, geography, and technology, all of which are closed-domain factual query tasks. The core focus is to assess the model's knowledge memorization and accurate matching capabilities without external tool assistance.

\textbf{SimpleQA \cite{simpleqa}}: A simple yet challenging benchmark for evaluating the factual accuracy of models. All question samples are fact-oriented and have verifiable answers.

\subsection{Reasoning}

\textbf{BBH \cite{bbh}}: A collection of reasoning tasks selected from the Big-Bench benchmark, covering 23 distinct reasoning scenarios (e.g., causal reasoning, analogical reasoning, logical deduction, symbolic reasoning, etc.). 

\textbf{HellaSwag \cite{hellaswag}}: A high-difficulty benchmark for evaluating commonsense natural language reasoning capabilities, containing 70,000 multiple-choice questions. Each question presents a specific scenario and four possible outcomes, tasking the model with selecting the most logically consistent conclusion.

\textbf{HLE \cite{hle}}: A high-difficulty benchmark at the forefront of human knowledge, designed to serve as the definitive closed academic benchmark in its class, covering a wide range of disciplines. It includes 2,500 questions across dozens of subjects, including mathematics, humanities, and natural sciences, co-developed by global subject matter experts, with both multiple-choice and short-answer question formats. OpenCompass supports its text-only subset.

\subsection{Computation}

\textbf{Competition-style Questions}: Datasets constructed based on original questions from international mathematics competitions, such as AIME, HMMT, and AMO-Bench.

\textbf{MATH \cite{math}}: A dataset containing 12,500 high-difficulty high school competition-level math problems, covering multiple domains from algebra to calculus, with detailed step-by-step solutions provided for each question.

\subsection{Science}

\textbf{PHYSICS \cite{physics}}: A high-level physics problem-solving benchmark designed to evaluate the reasoning and analytical capabilities of foundation models. It includes 1,297 doctoral qualifying examination questions across six fundamental physics disciplines. OpenCompass supports its text-only subset.

\textbf{ClimaQA \cite{climaqa}}: A benchmark for evaluating LLM performance on climate science question-answering tasks. Built on graduate-level climate science textbooks, it provides a reliable foundation for generating questions containing precise professional terminology and complex scientific theories.

\textbf{SmolInstruct \cite{smol}}: A large-scale, comprehensive, and high-quality domain-specific dataset for chemistry, focusing on small molecules as the core research object. It includes 14 carefully selected tasks and over 3 million samples.

\subsection{Language}

\textbf{MMMLU \cite{mmmlu}}: A multilingual version of the MMLU benchmark provided by OpenAI. It translates MMLU's test questions into 14 different languages through professional human translation, aiming to evaluate the model's multi-task understanding capabilities across different linguistic environments.

\textbf{PMMEval \cite{pmmeval}}: A multilingual benchmark covering both foundational knowledge assessment and specialized competence evaluation. It supports up to 10 languages from 8 language families, facilitating comprehensive evaluation of multilingual capabilities and comparative analysis of cross-linguistic transferability.

\subsection{Code}

\textbf{LiveCodeBench \cite{lcb}}: Provides a comprehensive and data contamination-free evaluation platform for large language model programming capabilities. Its questions are sourced from mainstream programming competition platforms such as LeetCode, AtCoder, and CodeForces, with core features including real-time updates and multi-dimensional programming skill assessment.

\textbf{BigCodeBench \cite{bigcodebench}}: A benchmark dedicated to evaluating LLMs' code generation capabilities in synthesizing programs from natural language descriptions or code snippets. It has also open-sourced a dedicated evaluation environment for detecting the quality of generated code.

\subsection{Long-text Processing}

\textbf{Ruler \cite{ruler}}: A benchmark specifically designed for long-context tasks. Extended from the traditional "needle-in-a-haystack (NIAH)" test, it not only assesses the model's ability to retrieve key information from lengthy distracting text but also features multiple configurable task settings, supporting flexible adjustments to sequence length and task complexity.

\textbf{LongBench \cite{longbench}}: A multi-task, Chinese-English bilingual long-text capability evaluation benchmark, comprising six categories and twenty-one distinct tasks. It covers key long-text application scenarios such as single-document QA, multi-document QA, summarization, few-shot learning, synthesis tasks, and code completion.

\subsection{Others}

\textbf{Arc-AGI \cite{arc}}: A dataset consisting of a series of logical rule-based reasoning tasks on a 2D coordinate system. Each task includes several "demonstration pairs" and one or more "test inputs," with the core goal of evaluating the model's general reasoning generalization and knowledge transfer capabilities.

\textbf{IFEval \cite{ifeval}} and \textbf{IFBench \cite{ifbench}}: Evaluate LLMs' instruction-following capabilities using a series of verifiable instructions. These instructions are characterized by being automatically and objectively verifiable for correct execution through simple programs.

%% file: sections/5.future.tex
\section{Future Works}

\textbf{Parallelization Enhancement.}
The current evaluation workflow of OpenCompass adopts a serial execution mode for Infer and Eval tasks: a corresponding Eval task can only be initiated after a single dataset completes all its full-scale Infer tasks, leaving room for scheduling optimization. In the future, we aim to maintain intra-dataset task seriality while achieving pipeline parallelism across multiple datasets. For instance, after Dataset A completes its Infer tasks and initiates the corresponding Eval tasks, Dataset B will be scheduled to start its Infer tasks simultaneously.

\textbf{Dialogue Template Expansion.}
Currently, OpenCompass only supports single-modal evaluation, which struggles to meet users' evaluation needs for multimodal perception and multi-turn context understanding capabilities in practical scenarios. In the future, we plan to expand the evaluable scope based on the ChatML format, enabling support for both multimodal evaluation and multi-turn dialogue evaluation.

\subsubsection*{Author Contributions}
The authors are listed in alphabetical order by their last names.

Maosong Cao, Kai Chen, Haodong Duan, Yixiao Fang, Zhiwei Fei, Tong Gao, Ge Jiaye, Mo Li, Hongwei Liu, Junnan Liu, Yuan Liu, Chengqi Lyu, Han Lyu, Ningsheng Ma, Zerun Ma, Yu Sun, Zhiyong Wu, Linchen Xiao, Zhuozhi Xiong, Jun Xu, Haochen Ye, Zhaohui Yu, Yike Yuan, Songyang Zhang, Yufeng Zhao, Fengzhe Zhou, Peiheng Zhou, Dongsheng Zhu, Lin Zhu, Jingming Zhuo

%% file: sections/6.appendix.tex
\section{Evaluation Results}

The baselines of each model on six general benchmarks, evaluated with OpenCompass, are presented in Table \ref{tab:model_performance}. The results are sourced from \url{https://rank.opencompass.org.cn/leaderboard-llm-academic/?m=REALTIME}

\begin{table}[htbp]
    \centering
    \caption{Model Performance Benchmarks}
    \label{tab:model_performance}
    \scriptsize
\begin{tabular}{c|ccccccc}
\hline
Models & Average & IFEval & HLE & \begin{tabular}[c]{@{}c@{}}GPQA\-\\ diamond\end{tabular} & \begin{tabular}[c]{@{}c@{}}AIME\\ 2025\end{tabular} & \begin{tabular}[c]{@{}c@{}}MMLU-\\ Pro\end{tabular} & \begin{tabular}[c]{@{}c@{}}LiveCode\\ BenchV6\end{tabular} \\ \hline
Gemini-3-Pro-Preview & 81.32 & 92.79 & 37.98 & 91.54 & 93.44 & 89.31 & 82.86 \\
GLM-5-FP8 & 78.98 & 93.16 & 28.13 & 85.35 & 95.83 & 85.23 & 86.19 \\
GPT-5-2025-08-07 (high) & 78.84 & 94.64 & 28.51 & 86.36 & 94.06 & 86.21 & 83.24 \\
DeepSeek-V3.2-Speciale & 78.22 & 91.68 & 28.55 & 86.74 & 96.04 & 85.45 & 80.86 \\
Kimi-K2.5 & 78.22 & 93.90 & 28.64 & 88.13 & 91.88 & 86.20 & 80.57 \\
GLM-4-7 & 77.61 & 90.20 & 25.39 & 86.87 & 95.42 & 83.97 & 83.81 \\
Step-3.5-Flash & 76.93 & 93.16 & 21.59 & 83.71 & 95.73 & 83.46 & 83.90 \\
Kimi-K2-Thinking & 75.33 & 92.42 & 21.31 & 82.70 & 94.06 & 84.34 & 77.14 \\
DeepSeek-V3.2 & 75.29 & 89.65 & 23.21 & 84.60 & 93.02 & 85.81 & 75.43 \\
Gemini-2.5-Pro & 73.62 & 90.02 & 21.12 & 84.72 & 88.65 & 85.85 & 71.33 \\
GLM-4.6 & 73.32 & 88.72 & 19.31 & 80.43 & 90.31 & 82.95 & 78.19 \\
Mimo-V2-Flash & 73.31 & 89.46 & 20.52 & 82.07 & 92.92 & 83.11 & 71.81 \\
MiniMax-M2.5 & 73.26 & 91.13 & 22.24 & 84.60 & 86.25 & 81.74 & 73.62 \\
gpt-oss-120b (high) & 73.16 & 90.20 & 18.34 & 78.91 & 93.44 & 79.68 & 78.38 \\
GPT-5-mini-2025-08-07 (high) & 73.15 & 92.79 & 21.49 & 82.70 & 90.94 & 70.24 & 80.76 \\
o3-2025-04-16 & 72.78 & 92.24 & 21.31 & 82.32 & 85.10 & 83.02 & 72.67 \\
Qwen3-235B-A22B-Thinking-2507 & 71.84 & 87.80 & 18.48 & 79.80 & 90.94 & 83.46 & 70.57 \\
TeleChat-Thinking & 71.79 & 94.27 & 20.29 & 81.44 & 78.96 & 84.09 & 71.71 \\
o4-mini-2025-04-16 (high) & 71.54 & 92.98 & 14.35 & 77.40 & 91.04 & 82.42 & 71.05 \\
o3-mini-2025-04-16 (high) & 70.20 & 91.87 & 10.82 & 79.42 & 83.85 & 79.24 & 76.00 \\
Qwen3-Next-80B-A3B-Thinking & 69.54 & 89.46 & 13.46 & 77.02 & 88.96 & 82.02 & 66.29 \\
MiniMax-M2 & 69.48 & 90.20 & 13.37 & 78.66 & 79.06 & 81.60 & 74.00 \\
GLM-4.5 & 69.25 & 85.40 & 16.95 & 79.55 & 85.83 & 82.72 & 65.05 \\
DeepSeek-R1-0528 & 68.07 & 80.04 & 14.39 & 80.56 & 88.96 & 83.52 & 60.95 \\
GPT-5-nano-2025-08-07 (high) & 67.23 & 90.20 & 10.96 & 67.80 & 85.42 & 75.02 & 74.00 \\
JT-Think & 67.18 & 94.64 & 14.44 & 77.65 & 65.94 & 85.57 & 64.86 \\
Seed-OSS-36B-Instruct & 66.45 & 83.73 & 9.52 & 71.46 & 86.04 & 80.62 & 67.33 \\
gpt-oss-20b (high) & 66.42 & 88.91 & 11.56 & 68.94 & 87.92 & 72.84 & 68.38 \\
Qwen3-30B-A3B-Thinking-2507 & 66.17 & 89.65 & 11.65 & 70.08 & 86.77 & 79.46 & 59.43 \\
doubao-seed-1.6-thinking-250615 & 66.00 & 78.93 & 13.46 & 75.88 & 82.81 & 83.11 & 61.81 \\
Qwen3-235B-A22B-Thinking & 65.83 & 85.77 & 12.67 & 73.23 & 82.40 & 80.89 & 60.00 \\
GLM-4.5-Air & 65.46 & 84.10 & 12.58 & 72.22 & 84.79 & 80.68 & 58.38 \\
Qwen3-Next-80B-A3B-Instruct & 62.50 & 87.62 & 8.03 & 74.12 & 69.17 & 81.31 & 54.76 \\
Finix-P1-32B (Thinking) & 62.01 & 88.72 & 6.92 & 68.69 & 70.21 & 82.65 & 54.86 \\
Claude Sonnet 4 (Thinking) & 61.82 & 88.35 & 8.73 & 74.62 & 68.65 & 83.02 & 47.52 \\
Qwen3-235B-A22B-Instruct-2507 & 61.30 & 88.35 & 12.30 & 75.51 & 69.48 & 79.22 & 42.95 \\
Qwen3-4B-Thinking-2507 & 60.60 & 88.54 & 6.04 & 64.65 & 80.00 & 72.78 & 51.62 \\
LongCat-Flash-Chat & 59.83 & 90.20 & 8.50 & 70.08 & 61.04 & 80.95 & 48.19 \\
MiniCPM-Sala & 58.68 & 76.89 & 9.29 & 58.21 & 81.04 & 71.99 & 54.67 \\
QwQ-32B & 58.16 & 81.52 & 7.75 & 64.02 & 68.96 & 73.94 & 52.76 \\
GLM-Z1-32B-0414 & 57.24 & 84.29 & 7.15 & 63.26 & 61.15 & 77.42 & 50.19 \\
Kimi-K2-Instruct & 57.04 & 88.54 & 6.73 & 72.73 & 49.06 & 78.39 & 46.76 \\
ERNIE-4.5-21B-A3B-Thinking & 56.98 & 81.15 & 6.55 & 60.23 & 76.15 & 70.79 & 47.05 \\
Qwen3-30B-A3B-Instruct-2507 & 55.26 & 83.92 & 7.66 & 63.64 & 63.75 & 73.93 & 38.67 \\
DeepSeek-Chat-V3-0324 & 54.81 & 81.89 & 5.57 & 67.05 & 45.10 & 83.27 & 46.00 \\
ERNIE-4.5-300B-A47B-PT & 54.67 & 88.91 & 3.81 & 82.95 & 32.60 & 78.72 & 41.05 \\
GPT-4.1-20250414 & 52.64 & 88.17 & 6.04 & 68.06 & 35.31 & 81.02 & 37.24 \\
Hunyuan-A13B-Instruct & 52.31 & 60.26 & 5.99 & 64.14 & 65.73 & 73.84 & 43.90 \\
Llama4-Maverick-17B-128E-Instruct & 48.25 & 87.06 & 5.34 & 69.07 & 15.21 & 79.56 & 33.24 \\
Qwen3-4B-Instruct-2507 & 47.20 & 82.44 & 5.06 & 52.27 & 46.88 & 63.03 & 33.52 \\
GLM-4-32B-0414 & 46.66 & 85.21 & 3.99 & 51.26 & 33.44 & 73.47 & 32.57 \\
Gemma-3-27B-it & 42.09 & 80.96 & 4.22 & 46.34 & 22.40 & 67.84 & 30.76 \\
\hline
\end{tabular}
\end{table}